\documentclass{article}
\usepackage{ijcai21}

\usepackage{times}
\usepackage{soul}
\usepackage{url}
\usepackage[hidelinks]{hyperref}
\usepackage[utf8]{inputenc}
\usepackage[small]{caption}
\usepackage{graphicx}
\usepackage{amsmath}
\usepackage{amsthm}
\usepackage{booktabs}
\usepackage{algorithm}
\usepackage{algorithmic}
\usepackage{amsmath,amsfonts,amssymb}

\newtheorem{definition}{Definition}
\def\reals{{\mathbb R}}

\title{Planning for Novelty: Width-Based Algorithms for Common Problems in Control, Planning and Reinforcement Learning~\footnote{A position paper for IJCAI 2021 Early Career Spotlight Talk}}
\author{
    Nir Lipovetzky
    \affiliations
    School of Computing and Information Systems,
    University of Melbourne, Australia
    \emails
    nir.lipovetzky@unimelb.edu.au
}

\begin{document}
\maketitle

\begin{abstract}
  Width-based algorithms search for solutions through a general definition of state novelty. These algorithms have been shown to result in state-of-the-art performance in classical planning, and have been successfully applied to model-based and model-free settings where the dynamics of the problem are given through simulation engines. Width-based algorithms performance is understood theoretically through the notion of \textit{planning width}, providing polynomial guarantees on their runtime and memory consumption. To facilitate synergies across research communities, this paper summarizes the area of width-based planning, and surveys current and future research directions.
\end{abstract}


\section{Introduction}

Planning plays a central role defining artificial and human intelligence \cite{simon:gps,kahneman2011thinking}. It is an active research area for different communities, each with their own set of assumptions, formulations and  computational approaches. The goal of this paper is to introduce one of the recent advances from the AI planning community, width-based planning \cite{nir:ecai12}, and illustrate its wider applicability to problem formulations common in other communities such as optimal control~\cite{bertsekas:dp} and reinforcement learning (RL)~\cite{sutton:book}. 

Classical planning deals with the problem of generating a sequence of decisions (plan) traversing a transition system while meeting a specification of the intended behaviour, expressed as a logical goal condition to be satisfied, or as a utility function to be maximized. The transition system of a planning problem is defined implicitly by a model which can vary along different dimensions. The main ones explored in this paper impose assumptions on 1) the uncertainty about the initial condition and the effects of each decision, 2) the dynamics represented by the model, 3) the language used to specify the intended behaviour, and 4) whether access to the implicit representation of the model is available.

Width-Based algorithms search in the transition system guided by a measure of novelty, defined as a function of the set of features $\Phi$ characterising each state of the system. The simplest set of features can be defined as an injective function mapping each state variable valuation into a unique feature. The novelty of a state is then measured in terms of the \textit{smallest subset of features} made true for the first time in the search. Note that this measure of novelty is agnostic on the model being solved, which lends itself to generalise well over different planning models. Remarkably, this measure alone suffices to define the complexity of a problem $\Pi$ in terms of its width $w(\Pi)$, independent of the initial and goal conditions of the problem, as well as proving the existence of width-based algorithms which run in time and space exponential in the width of the problem. These algorithms have been shown to yield state-of-the-art solvers over a wide range of problems. Width-based search departs from other blind search methods like Depth First Search and Breadth First Search, because it exploits the factorized structure of the states to guide its search.

\section{Novelty and Width-Based Search}

The main concept driving width-based search is the notion of state novelty~\cite{nir:ecai12}, which depends on states $S^-_s$ generated before a state $s$, and the set of features $\Phi(s) = \{f \ |\  f\in s\}$ true in each state.

\begin{definition}
The novelty of a newly generated state $w(s) =  \min_{\phi \subseteq \Phi(s), \phi \not\subseteq \Phi(s'), s'\in S^-_s} |\phi|$ is the size of the smallest subset of features true in the state, and false in all previously generated states $S^-_s$. 
\end{definition}

In classical planning, without loss of generality states are described in terms of boolean fluents $F$, it is natural then to define the state features as $\Phi(s) = \ \{ f \ |\  f \in s, f \in F\}$. Therefore, the novelty of states in classical planning ranges from 1 to $|F|$, given that the smallest subset of features contains a single fluent, while the largest contains all the fluents of the problem. The number of subsets, and hence the number of possible novel states is $O(|\mathcal{P}(F)|)$. 

The simplest width-based search algorithm exploiting novelty is \textit{Iterative Width}. 
IW($k$) is a breadth-first search (BrFS) pruning states whose novelty is greater than $k$. IW($k$) considers features up to size $k$ and runs in time and space exponential in $k$, i.e.  $O(|F|^k)$. Remarkably, in spite of IW being incomplete,~\citeauthor{nir:ecai12}~\shortcite{nir:ecai12} showed that the linear and quadratic time IW(1)  and IW(2) solve respectively 37\% and 88\% of 37,921 instances from benchmarks up to 2012, where each problem with a goal made of $|G|$ fluents is split into $|G|$ problems, each with a single goal fluent. This performance is not accidental, as it is possible to prove that any instance of most planning domains can be solved by IW(2) provided that the goal $|G|=1$ is a single fluent.

The key notion exploited by IW, and many subsequent width-based algorithms is the notion of \textit{problem width}~\cite{nir:ecai12}. Similar \textit{treewidth} notions have been proposed to  bound algorithms over intractable problems in constraint satisfaction  and bayesian networks~\cite{dechter:book,pearl:book}. 

\begin{definition}
The width $w(P)$ of a classical planning problem $P$ is the minimum $k$ for which there is a sequence $\vec{\Phi} = ( \phi_0, \ldots, \phi_n )$ of feature subsets, each $|\phi_i|\leq k$ with at most $k$ features, such that 1) $\phi_0 \subseteq \Phi(s_0)$ is true in the initial state, 2) \textbf{any optimal plan} achieving $\phi_i$ can be \textbf{extended} into an optimal plan for $\phi_{i+1}$ with a \textbf{single action}, $i=0,\ldots,n-1$, 3) any optimal for $\phi_n$ is an optimal plan for the problem $P$. 
\end{definition}

If the goal is true in the initial state, for convenience we set $w(P)=0$, and $w(p) = |F|+1$ if the problem is unsolvable. Note that there is no assumption imposed upon how the transition function is specified, the definition only assumes a factored representation of the states in terms of state features. Intuitively, $\vec{\Phi}$ defines a sequence of sub-goals that constraints the search to ensure the goal will be reached optimally. Width characterises the size of the biggest constraint needed. If the width of a problem is known to be $w(P)=k$, then IW($k$) is guaranteed to solve the problem optimally (shortest path)~\footnote{If the problem has non-uniform costs, in order to guarantee optimality IW should run Dijkstra instead of BrFS.}.~\citeauthor{nir:book}~\shortcite{nir:book} has constructive proofs for several known domains such as Blocks world, Logistics, and 15-puzzle, showing that independently of their size and initial situation, each single goal fluent can be reached with width 2. When the width of the problem is unknown, IW runs by iteratively increasing the bound from 1 to $|F|$ until the problem is solved. IW($|F|$) is equivalent to BrFS pruning only true state duplicates, if no solution is found, then the problem is unsolvable. While IW is not practical for planning with larger goals $|G|>1$, it has been shown to work well for Finite Horizon Optimal Control and RL problems (Section~\ref{sec:numeric}), and it underpins the ideas in other state-of-the-art planners for satisficing planning known as best first width search (BFWS).

State-of-the art classical planners rely on best first search algorithms (BFS) guided by goal-oriented heuristics derived automatically from the specification of the problem ~\cite{geffner2013concise}. One of the known pitfalls of BFS is that it can get trapped in large plateaus. To avoid this, BFWS ~\cite{nir:aaai2017} balances  exploration and exploitation by integrating novelty with goal directed heuristics 
through the evaluation function $f = \langle w_{H},\vec{h} \rangle$, where $\vec{h}$ is an ordered sequence of heuristics, and $w_{H}$ is the novelty measure. Given a state $s$, novelty in BFWS is computed with respect to states $s' \in S^-_s$ seen before with the same heuristic value $h(s) = h(s')$, for all $h \in H$. $H$ is a set of heuristics that partition the state space for a more refined computation of novelty. The evaluation function $f$ breaks ties lexicographically, preferring novel states first, and then breaking ties by goal-oriented heuristics. Preferring first the exploration term over the exploitation term is beneficial, as it limits the search effort spent escaping heuristic value plateaus. 

One of the best performing BFWS variant uses a single  heuristic $\vec{h}=(\#g)$ counting the number of unachieved goals, and novelty uses $H=\langle \#g, \#r \rangle$ the goal counting heuristic along with a counter $\#r$ keeping track of how many relevant fluents $R\subseteq F$ have been achieved along the path to the current state $s$ from the initial state. The relevant set $R$ is defined as the fluents that appear in a polynomial relaxation known as delete-relaxed plan~\cite{geffner2013concise}. 
 This version won the agile track of the last international Planning Competition (IPC)~\cite{nir:ipc18}, which measures the speed and number of solutions found in 5 minutes. An incomplete but polynomial version that instead uses the novelty measure to prune states with novelty $W_H>1$, solved 73\% of the problems with $|G|>1$, independent of the size of their goal. This polynomial version, 1-BFWS, can quickly solve or return no solution, and it can be used as a preprocessing step to find a solution. This was instrumental for DUAL-BFWS second place on the satisficing track, where 1-BFWS  run first, and a complete BFWS incorporating other planning heuristics was run if no solution was found by 1-BFWS. In fact, it was shown that a family of polynomial $k$-BFWS algorithms can render close to state-of-the-art performance with provable bounds~\cite{nir:icaps2017}. BFWS  has performed well beyond classical planning. It has been used for decentralized multi-agent planning ~\cite{nir:icaps19}, and regression for classical planning~\cite{lei2021width}.
 Other notable width-based planners have shown great performance with an alternative novelty formulation~\cite{katz2017adapting} and integrated in Enforced Hill Climbing~\cite{fickert2018making} and BFS with lookaheads~\cite{fickert2020novel}.

\section{Planning with Simulators}

A model described formally by a planning language is a key requirement to enable the derivation of informed heuristics. Having access to the structure of the actions and state variables resulted in an active area of research, improving performance of existing solvers through several relaxations underpinning existing heuristics ~\cite{geffner2013concise}. The language role was not only descriptive, but also computational. Yet, it is known that every language has its expressive limitations, some problems are easier to model than others. Take for instance Pacman, the behaviours of the ghosts are easily specified pragmatically but not declaratively~\cite{frances2017purely}. Other problems are just easier to specify by non-planning experts via simulation engines. Take the Atari games as an example, or platforms like OpenAI-Gym used to specify RL problems
. The downside of using simulators to specify classical planning problems is that most solvers will not work, as their heuristics require access to the specification of the problem.

\citeauthor{frances2017purely}~\shortcite{frances2017purely} insight was to realize that state-of-the-art BFWS algorithms do not require access to the transition function, novelty and goal counting only rely on a factored representation of the states. The only term that requires access to the action' structure is the $\#r$ counter, as the set of relevant fluents are computed by the delete relaxation. Hence, \citeauthor{frances2017purely}~\shortcite{frances2017purely} proposed alternative methods to derive the relevant fluents over problems specified via simulators. Namely, the linear IW(1) or quadratic IW(2) were used to create a lookahead at the initial state, and fluents on the way to paths achieving any single goal were marked as relevant. The experimental results over the set of problems from IPC-2014 showed that this version, which does not have access to the structure of the transition function and does not impose a restriction over how the model is specified, performed as well as other planners that had access to the model. These results were a strong argument towards using planning solvers beyond the scope defined by a language, and a bridge towards other research areas whose techniques focus on problems specified through simulators (model-free). In fact, this enabled BFWS to be used over theories encoded via external reasoners for Task and Motion Planning~\cite{ferrer2017combined}, track beliefs online over POMDPs for transparent planning~\cite{macnally2018action}, and to reason over epistemic knowledge while planning~\cite{hu2019you}. 

\section{Numeric Features}
\label{sec:numeric}
So far, the discussion of width-based algorithms has focused on classical planning, as their performance is understood in terms width $w(P)$. This section discusses how these algorithms have been used beyond classical planning, but, without provable performance guarantees, given that a width notion for these problems is yet to be discovered. 

\noindent\textbf{Classic Control Problems}.
Classic deterministic control problems differ from classical planning mainly in three dimensions: 1) state variables are numeric, 2) continuous or discrete dynamics are typically encoded via a simulator, 3) and the target behaviour is often expressed via a cost (reward) function to be maximized (minimized). None of these changes preclude the use of width-based algorithms, as long as state features are defined over numeric state variables. 

\citeauthor{ramirez2018integrated}~\shortcite{ramirez2018integrated} showed that complex non-lineal aircraft simulations can be controlled by running IW($1$). In this context, IW($1$) can be understood as an approximate dynamic programming algorithm, solving the model predictive control problem iteratively over a rolling horizon window~\cite{bertsekas:dp}. Each problem is a finite horizon optimal control problem which IW(1) solves, without optimality guarantees. IW(1) builds a lookahead by expanding novel states only, and returning the action leading to the state with the least accumulated cost. The features used for novelty map each state variable $x_k^i \in \reals$ valuation into a unique feature $f_x^i: x^i \mapsto [0, 2^B -1]$, where $i$ is the \textit{i-th} state variable, and $B$ is the number of bits in the floating point representation, effectively, the binary encoding of the number. This definition of features is finite, given that existing models of computation only allow for finite arithmetic precision. IW(1)  maps a state into its features $\Phi(x_k) = \ \{ f_x^i \ | x^i \in x_k \}$, only expanding  states with a variable whose real value has not been seen before. This search, which ignores completely the objective of the optimization problem, is powerful enough to create  real-time aircraft manoeuvres from  first principles that follow dynamical constraints and the cost function.

In tasks with sparse rewards or with rewards attainable only if the goal is reached and maintained until a given horizon, these naive novelty features do not work well.~\citeauthor{ijcai2020}~\shortcite{ijcai2020} proposed an alternative general dynamic encoding of state features taking into account the dynamics of the problem, and the choices made by the width-based search algorithm. These features, known as Boundary Extension Encoding (BEE) features, are designed to mark as novel those states that push the \textit{boundaries} of a state variable, keeping track of the valuations already discovered by the search. If a state variable extends the boundaries, a new interval with a new unique index is created tracking the range of the extension, otherwise, the index of the interval where the state variable belongs is returned. Novelty is then computed by mapping a state into its BEE interval index for each state variable. This encoding is used by polynomial versions of $k$-BFWS with no heuristic information, where novelty alone outperforms Deep RL over classical control problems. It can solve Mountain Car in 4.6 seconds whereas PPO2~\cite{schulman2017proximal} cannot learn a valid policy after 7 hours. In Acrobot, PPO2 takes 20 minutes longer than $k$-BFWS to find a policy of the same quality. In CartPole, $k$-BFWS finds the optimal solution stabilizing the pole across the full trajectory, whereas PPO2 takes 3 times longer to find such policy. If an alternative IW version known as Rollout-IW~\cite{bandres2018planning} is used, a decision limit of 200ms suffices to solve the problem in real time. Compared with LQR controllers
, k-BFWS finds solutions of similar quality, but it is significantly better over Acrobot when initial states are perturbed, as this breaks the controllability guarantees of the LQR solution.

\noindent\textbf{Reinforcement Learning Problems}.
Deterministic RL problems differ from classical planning in the same dimensions as control problems, the transitions are encoded through a simulator, a goal region is absent but instead rewards are used, where the optimal solution aims to maximize the accumulated reward over a fixed horizon.

The Arcade Learning Environment is a popular simulator for testing RL algorithms~\cite{bellemare2013arcade}, supporting 54 Atari games with a horizon of 18,000 steps, amounting for 5 minutes of real-time gameplay. RL approaches have been successfully applied over this benchmark
. An alternative approach uses a approximate dynamic programming algorithms like IW over a fixed budget to build a lookahead tree, returning the best action, and moving to the next step. IW($1$) was shown to outperform UCT, a Monte Carlo Tree Search algorithm akin to the one used in the solver that outperformed Lee Sedol in Go
, using each byte of the 128 Bytes RAM as a novelty state features  $f_i\in[0,255]$, $i=1,\ldots,128$~\cite{nir:ijcai15}. That is, a state would be novel if it has a byte value which has not been seen before. This measure allowed for the search to reach far away rewards, performing better than UCT over 31 out of 54 games, whereas UCT outperformed IW(1) in 19 games. IW(1) results were further improved by breaking ties in the expansion of states with the accumulated reward, and refining the definition of novelty by marking a feature as novel if it was achieved by a state with a larger accumulated reward than the best state seen so far with that feature~\cite{shleyfman2016blind}. This sufficed for IW to outperform UCT in 44 out of 53 games. Furthermore, when screen features capturing spacial and temporal relations of pixels taken from the screen input state are used to define novelty, along with  RIW, a width-based search that uses depth-first rollouts instead of BrFS as the underlying search, then, RIW with a real-time budget of 0.5 seconds is the best in 30,6\% of the games, while Deep RL
, Shallow RL
, and a human expert are best in 24.4\%, 12.2\% and 32.6\% respectively. With a higher budget of 32s per action RIW outperforms human experts in 75.5\% of the games~\cite{bandres2018planning}.

\section{Research Directions}

\paragraph{Novelty Approximations for High Width.}
Checking that a state has novelty $k$ requires in the worst case keeping track in memory of all previously seen $|\Phi|^k$ features, as well as enumerating the same amount of features per state. This computation makes it impractical to compute novelty values greater than 2. Recently,~\citeauthor{singh2021approximate}~\shortcite{singh2021approximate} have shown that sampling techniques can be used to enumerate state features, and Bloom filters can be used to track the features seen so far, bringing the complexity down to linear time, i.e. $O(k|\Phi|)$. This novelty approximation can increase or decrease the width of the problems, but this error probability has been characterized and shown that in practice is quite low. Furthermore, ~\citeauthor{singh2021approximate}~\shortcite{singh2021approximate} proposed an approximation of BFWS using an adaptive policy to forego the expansion of some states, controlling the exponential growth of states with novelty $k$, ensuring that states with high novelty can be expanded. This algorithm manages to exploit a higher range of novelty values, and is faster than other state-of-the-art Width-based planner. This line of research can enable solving high width problems with polynomial guarantees.

\paragraph{Serializations and Width.}
\citeauthor{nir:book}~\shortcite{nir:book} suggested that classical planning problems are hard to solve if they either have high atomic width, or their goals $G$ are hard to serialize into a sequence of sub-problems, achieving one subgoal at a time. A simple serialized IW (SIW) algorithm doing hill-climbing over the set of goals, using IW to achieve one more single goal $g\in G$ at a time until all goals are achieved, was sufficient to show that 71\% of problems with $|G|>1$ were solvable with low width, once this serialization was attempted. ~\citeauthor{frances2017purely}~\shortcite{frances2017purely} showed that novelty features can be encoded programmatically as control knowledge to lower the width of a problem and help such serializations. This begs the question on whether such serializations can be encoded via domain knowledge or through state novelty features in such a way that problems with $|G|>1$ have provable width $w(P)=1$. If such encoding can be generated automatically, planning problems could be solved with linear time guarantees with no restriction on the size of their goal. Recently,~\citeauthor{bonet2021general}~\shortcite{bonet2021general} proved the connection between generalized planning and planning width. Generalized plans take the form of a policy mapping state features into actions and solve multiple instances of a problem at once. The same language used to specified the rules of a general policy can be used to specify general serializations for planning problems, and their performance can also be understood in terms of their planning width.  ~\citeauthor{drexler2021expressing}~\shortcite{drexler2021expressing} used this language to elaborate hand crafted serializations encoded as policy sketches to guide SIW with provable low polynomial bounds. Known problems accepting polynomial suboptimal solutions such as Floortile, Barman, Childsnack, Driverlog have width $w(P)=1$ and Grid, Tpp, and Schedule have width $w(P)=2$ once a suitable serialization  that works for all possible goals and initial states is provided. This line of research has the potential to enable the derivation of such serializations automatically, solving large family of problems with provable performance bounds.   

\paragraph{Integration of Novelty Planning and Learning.}
How to best integrate novelty planning and learning is an active research area.~\citeauthor{junyent2019deep}~\shortcite{junyent2019deep} showed over the Atari problems that a policy can be learnt to guide IW into promising search areas, and the last hidden layer of the neural network can be used to encode novelty features, improving the performance of IW further. This suggests that suitable feature encodings can be learnt to improve width-based planners in classical planning, as well as learning goal serializations~\cite{junyent2021hierarchical}. IW planners can also benefit learning algorithms if they are used to bootstrap the learning process, e.g., finding a good first solution for problems like Mountain Car, and improve their sample efficiency. These research lines can benefit from synergies in both fields.

\paragraph{Novelty and Pseudo-Counts.}
Pseudo-counts~\cite{bellemare2016unifying} were introduced as an exploration bonus for RL algorithms, but their connection to novelty and width has not been studied yet. Novelty specific features have also been widely adopted over evolutionary algorithms, introduced by~\citeauthor{lehman2011abandoning}~\shortcite{lehman2011abandoning}, but their performance have not been connected with a similar planning width bound. If these connections are formalized, alternative exploration methods can be discovered for planning problems, novelty measures with provable bounds may be formulated over new classes of problems, along with new polynomial algorithms.

\paragraph{Tractable fragments beyond Planning.}
Planning width has been formulated over classical planning problems, explaining the good performance of width-based algorithms, yet, these algorithms have been shown to perform well over harder problems with a stochastic transition function~\cite{geffner2015width,nir:2019width}. These problems are characterized as MDPs, the formal model underlying most RL problems. What remains open is to prove that a similar width parameter over MDPs exists. This may lead to new algorithms tapping into the proof's insights.





\section*{Acknowledgments}
I thank IJCAI 2021 program committee for the invitation to speak at the early-career spotlight, my former Ph.D. supervisor, current collaborators, and the planning community for the opportunity to advance this research together.
\bibliographystyle{named}
\bibliography{control}

\end{document}